\pdfoutput=1

\documentclass[11pt]{article}

\usepackage[]{acl}

\usepackage{times}
\usepackage{latexsym}
\usepackage{graphicx} 
\usepackage{booktabs} 
\usepackage{lipsum} 
\usepackage{stfloats}
\usepackage{tabularx}
\usepackage{soul}
\usepackage{adjustbox}
\usepackage{appendix}
\usepackage{ctable}
\usepackage{amssymb}
\usepackage{mathtools, nccmath}
\usepackage{amsmath}
\usepackage{stmaryrd}  
\usepackage{caption}
\usepackage{subcaption}
\usepackage{multirow}
\usepackage{epstopdf}
\usepackage{paralist}
\usepackage[utf8]{inputenc}
\usepackage{multirow}
\usepackage{xstring}
\usepackage[T1]{fontenc}

\usepackage[utf8]{inputenc}

\usepackage{microtype}

\usepackage{inconsolata}

\title{The Emotion Dynamics of Literary Novels}

\author{Krishnapriya Vishnubhotla$^{1,3}$ \and Adam Hammond$^{2}$ \and Graeme Hirst$^{1}$ \\
        \textsuperscript{1}Department of Computer Science, University of Toronto \\ \textsuperscript{2}Department of English, University of Toronto \\ \textsuperscript{3}Vector Institute for Artificial Intelligence \\ \texttt{vkpriya@cs.toronto.edu adam.hammond@mail.utoronto.ca gh@cs.toronto.edu}
        \AND
        Saif M. Mohammad \\ National Research Council Canada  \\ \texttt{saif.mohammad@nrc-cnrc.gc.ca}
        }
        
\begin{document}
\maketitle
\begin{abstract}
Stories are rich in the emotions they exhibit in their narratives and evoke in the readers. The emotional journeys of the various characters within a story are central to their appeal. Computational analysis of the emotions of novels, however, has rarely examined the variation in the emotional trajectories of the different characters within them, instead considering the entire novel to represent a single story arc. In this work, we use character dialogue to distinguish between the emotion arcs of the narration and the various characters. We analyze the emotion arcs of the various characters in a dataset of English literary novels using the framework of Utterance Emotion Dynamics. Our findings show that the narration and the dialogue largely express disparate emotions through the course of a novel, and that the commonalities or differences in the emotional arcs of stories are more accurately captured by those associated with individual characters. 
\end{abstract}

\section{Introduction}
Storytelling is one of the earliest of all human traditions, predating even the invention of writing. Emotions are a powerful aspect of storytelling in any form, be it the oral tradition, or with the written word in fictional novels: a compelling story is often one that is able to evoke a strong emotional reaction from its audience. Characters (fictional named entities) are often at the center of this emotional connection that readers form with a story; we identify with their struggles, react to their pain, and celebrate their victories. The different characters within a single story usually display varying emotional trajectories through the course of the narrative. 
This variation is also seen at the level of the novel itself: some are tragedies, some are comedies; some have very dramatic highs and lows, others have a more even emotional tone. 

Kurt Vonnegut, in his now-famous lecture\footnote{\url{https://bigthink.com/high-culture/vonnegut-shapes/}}\cite{vonnegut2009palm}, spoke about the ``shapes of stories", plotting the fluctuations of a \textit{character's good or bad fortune} on the y-axis as a function of narrative time, from beginning to end, on the x-axis. One can assume that he was referring to the emotional journey of the protagonist of the story in his analysis. Prior work in NLP on plotting the shapes of stories, however, has considered a novel as representing a single emotional trajectory, where the dialogue of all the different characters is merged together with the narration to create one overarching novel arc (a simplification made because of the lack of annotated data in which characters are mapped to their utterances --- still a chaNational Research Council CanadaNational Research Council Canadallenging NLP problem).
In this work, we look instead at the emotion arcs of the \textit{individual characters} in  story, distinguishing them from the arcs of the narration as well as the overall novel arc, and quantify the extent of their variation both within a single novel and across novels written by different authors. 

Our goal is to quantitatively capture  \textit{longitudinal patterns} of a character’s emotional states (how emotions change over time), referred to as Utterance Emotion Dynamics (UED) \citep{hipson2021emotion}. Through these metrics we examine the following research questions about emotion arcs and emotion change (at an aggregate level) in full-length English novels:\\[-18pt]

\begin{enumerate}
\item \textit{The Emotion Dynamics of Novels}: How does the emotion change from the start of the novel to the end: overall, for just the narration, and for individual characters?\\[-20pt]
\item \textit{Narration vs.\@ Dialogue}: Do narration and dialogue have distinct emotion arcs?
 Prior work on the shapes of stories has largely glossed over the differences between characters, and considered either the entire novel to be a single trajectory, or worked with the text of the narration alone. We examine the validity of this approach by quantifying the differences between the emotions expressed in the narration and the dialogue of various characters.\\[-20pt]
\item \textit{Diversity of Character Arcs}: How diverse are the emotion arcs of the characters in a story?
The variety in the emotional trajectories of characters within a novel can be quite informative about the type of story it tells. A typical hero-versus-villain story, for example, might display opposing emotional arcs for the two main characters; one that follows the adventures of a group of friends might have more closely-correlated character arcs.\\[-20pt] 
\item \textit{Effects of Gender}: Are there consistent differences between the emotion arcs of gendered character groups? How does this change with different authors?\\[-20pt] 

\end{enumerate}

To answer these questions we conduct experiments that
make use of the following resources:\\[-16pt]
\begin{enumerate}
\item The \textit{Project Dialogism Novel Corpus (PDNC)} \cite{vishnubhotla-etal-2022-project}, which contains manual annotations identifying the speakers of all dialogue in 28 full-length English-language novels.\\[-20pt]
\item The NRC Valence, Arousal, and Dominance Lexicon \cite{vad-acl2018}, which includes $\sim$20,000 English words with a real-valued association scores (between 0 and 1) for the three dimensions of valence (V), arousal (A), and dominance (D) dimensions. A score of 1 indicates a maximum association or highest V/A/D, and a score of 0 indicates that the word is associated with the lowest V/A/D.\\[-20pt]
\item A simple, accurate, and interpretable way to generate emotion arcs from sequential text using an emotion lexicon \cite{Teodorescu2023EvaluatingEA}.\\[-20pt]
\item Metrics of Emotion Dynamics \cite{Hollenstein2015ThisTI, KUPPENS201722}, 
which quantify patterns of emotion change.\\[-20pt]
\end{enumerate}
We find that novels, on average, express emotions high in valence and lower in arousal and dominance (0.65 vs 0.38 and 0.52). Most of the high valence and dominance is expressed in character dialogue rather than narration. We show that emotional arcs of characters are quite different from that of the narration, and from one another (average correlations close to 0); the extent of this variation also changes from novel to novel. 
We find that female authors write characters who express higher valence, lower arousal, and higher dominance; male characters written by male authors have the highest arousal for their utterances.

Our work sheds light on aspects of storytelling that have been under-explored in computational literary studies, and quantitatively demonstrates the importance of centering characters in order to gain a more nuanced understanding of novels. We make our data and code publicly available.\footnote{\url{https://github.com/Priya22/literary-emotion-dynamics}}

\section{Background and Related Work}



Large-scale analyses of the flow of emotion-bearing words in novels as a whole has been explored since the early 2010s \cite{Mohammad2012FromOU,Reagan2016TheEA,Kim2017InvestigatingTR}.\footnote{Several resources and tools have also been developed for this such as the NRC Emotion Lexicon, Syuzhet, LIWC, VADER, and SentiWordNet.}
Work on individual character dialogues has been much sparser.
\citet{Nalisnick2013CharactertoCharacterSA} dive into emotions conveyed in character dialogue with a dataset of Shakespeare's plays; an utterance is assumed to have been uttered by the closest character mentioned. They use this to test various hypotheses on the relations between characters and their evolution, such as protagonist--antagonist pairs. 


\textit{Emotion dynamics} is a framework from psychology for measuring how an individual’s emotional state changes over time \cite{Hollenstein2015ThisTI, KUPPENS201722}. It has been studied in relation to emotional, mental, and physical well-being. \citet{hipson2021emotion} introduce a computational framework to measure emotion dynamics metrics from the emotion arc of an individual's natural language utterances over a time period. They term this framework \textit{Utterance Emotion Dynamics (UED)}. The authors primarily characterize an emotion state as a point in the two-dimensional valence--arousal space, but state that it can be over one or more emotion dimensions. In the present work, we limit ourselves to analyzing emotions individually, in a one-dimensional space, where the temporal flow is represented on the x-axis and the emotion state values on the y-axis; this follows the work done by \citet{VM2022-TED}, who developed UED metrics further for the one-dimensional case. 
\begin{table}[]
\small
    \centering
    \begin{tabular}{ll|rr}
      \textbf{Speaker Type} & & \textbf{Count} & \textbf{\#tokens} \\
     \hline
        Meta & novel & 28 & 96973.96 \\
        & narration & 28 & 57995.43 \\
        Characters & major & 111 & 6547.91 \\
        & intermediate & 113 & 2025.35 \\
        & minor & 585 & 231.99 \\
        \hline
    \end{tabular}
    \vspace{-1mm}
    \caption{The number of speakers, and average number of tokens, for each group of speakers that we consider, including meta-speakers.}
    \label{tab:basic-stats}
    \vspace{-5mm}
\end{table}

Given a sequence of temporally ordered utterances, the emotion state at a time point is defined as the average emotion value of a small window of utterances (or words) uttered around that time point. 
This window is moved forward by one word at each step to obtain a sequence of temporally ordered emotion states (the overlapping windows lead to a smoother and more continuous arc, when compared to using non-overlapping adjacent windows). The \textit{home base} for an individual is the space of emotion states, or values, that they are most likely to be found in, and is captured the \textit{mean} (average emotion state value), which is the center of the home base, and the \textit{variability} (standard deviation of the emotion states), which defines the bounds on either side. The home base can be visualized as an ellipse in the 2D space; for a single dimension, it will define a range of numerical emotion state values. 
Any motion of the emotion state outside of the range of this home base is termed a \textit{displacement}. Displacements in turn are characterized by a series of metrics: their temporal length (\textit{displacement length}), the peak emotional displacement in relation to the home base (\textit{peak distance}), and the \textit{rate} of rise to the peak and return to the home base. We describe the various UED metrics in more detail in Appendix \ref{appendix-a}.

The emotion value of a window of words can be determined in many ways: with lexicons that associate words and phrases with a numerical score along a particular emotion dimension, or with statistical and neural models that are trained to predict an emotion score given a text span as input. Lexicon-based methods are simpler and more interpretable, and do not need domain-specific training or fine-tuning of models. \citet{Teodorescu2023EvaluatingEA} also showed that lexicon-based methods were able to match the true emotion arc of a text sequence with a high accuracy (\textbf{correlations above 0.9}) when using window sizes of a 100 words or more. Emotion lexicons in turn have been created for many emotion dimensions, and in multiple languages (LIWC \cite{Tausczik2010ThePM}; WordNet-Affect \cite{Bobicev2010EmotionsIW}, SentiWordNet \cite{baccianella-etal-2010-sentiwordnet}, VADER \cite{Hutto2014VADERAP}, and the NRC suite of emotion and affect lexicons \cite{LREC18-AIL, vad-acl2018}). We therefore follow this approach in our work.
\section{Literary Emotion Dynamics}
There are several emotional trajectories of interest given a novel: one can look at the emotions of entire novel text, as has been done in prior NLP work, or consider the narration and dialogue as two text streams of interest, representing the narrator and the characters (we term these text streams as being uttered by \textit{meta-speakers}). The dialogue can further be analyzed by considering each character's utterances as an individual text stream of interest. 

We categorize the characters in a novel into three groups based on the volume of their dialogue: \textit{major characters} are those who contribute at least 10\% of the total dialogue in the novel or have at least 100 attributed quotations; \textit{minor characters} utter fewer than 35 quotations throughout; the rest are labelled \textit{intermediate}. Table \ref{tab:basic-stats} presents the number of speakers and the average number of tokens for each of these speaker groups.

The temporal ordering of the text in a story can in turn be viewed according to narrative time or chronological time. We order utterances by narrative time, which is of substantial interest for literary analysis as it is the order in which the reader experiences the text uttered by a character\footnote{Determining chronological time in a narrative text is a separate research problem on its own.}. We normalize the time for each speaker to lie between the range $[0,1]$, i.e, each speaker starts their emotion state at time point 0 and ends at time point 1, irrespective of the volume of their dialogue.\footnote{One could also conduct analyses of relative positions of when a character speaks within a novel; however, we leave that for future work. This paper focuses on patterns of change \textit{within} the emotion arc of a character.} 


We computed the emotion arc for a speaker 
using the parameters shown to produce high-quality emotion arcs by \citet{Teodorescu2023EvaluatingEA}: a rolling window size of 500 words, with the window moving forward by one word at each step until the final window subsumes the final word of their utterances.
In order to compute the emotion state at a particular timepoint (i.e, for a window of words), we use the NRC-VAD lexicon \cite{vad-acl2018}.

\begin{table*}[]
{\small
    \centering
    \begin{tabular}{l|rr|rr|rr|}
      \textbf{Speaker} & \multicolumn{2}{c|}{\textbf{Valence}} & \multicolumn{2}{c|}{\textbf{Arousal}} & \multicolumn{2}{c|}{\textbf{Dominance}} \\
      \hline
       &\textbf{Mean}&\textbf{Var.} &\textbf{Mean}&\textbf{Var.} &\textbf{Mean}&\textbf{Var.}\\
      \hline
    Novel (meta)&0.6472&0.0531&0.3811&0.0466&0.5192&0.0582\\
    Narration (meta)&0.6273&0.0567&0.3857&0.0462&0.5076&0.0603\\
    Character&0.6746&0.0295&0.3766&0.0282&0.5380&0.0355\\
    \hline
    \end{tabular}
    \caption{Aggregated Mean and variability (Var.\@) scores for the different types of speakers in our dataset, for each of the three emotion dimensions.}
    \label{tab:agg-avg}
    }
     \vspace*{-4mm}
\end{table*}

\subsection{Aspects of Gender}
For any individual, gender is a complex identity that can vary based on their personal preferences, experiences, and social expectations. \citet{cao2020toward} describe the various facets of gender experienced by humans, and their expression in linguistic form. For the characters in PDNC, we manually annotate gender information by primarily considering the referential pronouns used for them throughout the text, along with lexical indicators like highly-gendered names and nouns (for example, identifying \textit{Mary Jane} as Female, and \textit{the policeman} as Male). While gender can be expressed along a spectrum of identities, we do not find many characters in the dataset outside of the male/female binary (the exceptions are a few minor animal characters in \textit{Alice's Adventures in Wonderland}). The same is true of the authors represented in PDNC --- we rely on descriptions from Wikipedia to determine their presented gender, and do not find any who fall outside of the male/female binary (details in Appendix \ref{appendix-pdnc}). Our analysis therefore is restricted to these two categories.

\section{Research Questions and Experiments}

We now explore a series of research questions (RQs) on the emotional dynamics of literary novels, the diversity of emotion arcs that can be found within a novel, as well as across novels, and how they relate to one another. 

\subsection{The Emotion Dynamics of Novels (RQ1)}

\begin{figure*}
	     \centering
	     \includegraphics[width=\textwidth]{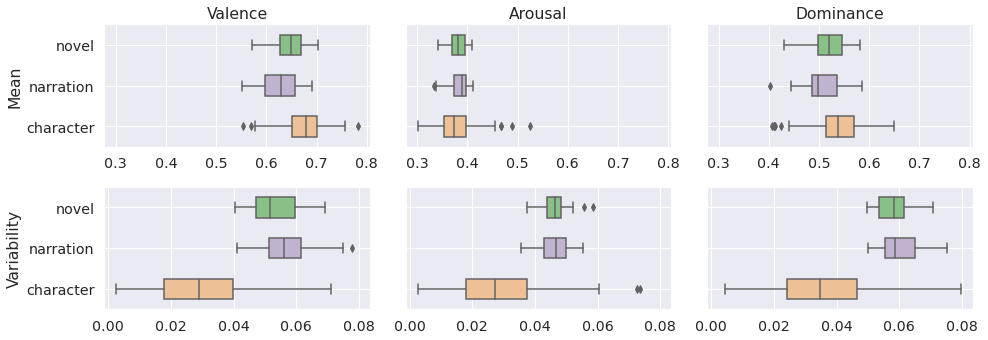}
	     \caption{Boxplots of the distributions of mean and variability for each of the three affective dimensions and all novels in PDNC, when the entire novel and the narration are considered to be uttered by a single meta-speaker, and when each character's utterances are considered individually.}
	     \label{fig:vad-mean-std-meta-dist}
       \vspace*{-4mm}
	 \end{figure*}

We generated emotion arcs and 
computed UED metrics for all the novels in PDNC. 
Figure \ref{fig:vad-mean-std-meta-dist} 
shows the distribution of the mean and variability for 
the entire novel, the narration, and individual characters. Table \ref{tab:agg-avg} lists the aggregated scores for mean and variability.
All other UED metrics are reported in Tables \ref{tab:valence-ued-all}-\ref{tab:dom-ued-all} in Appendix \ref{appendix-ued-metrics}.

It is immediately apparent that there is much more variation in the emotion dynamics of individual characters than when the narrative and/or the entire novel are clumped together into a single voice. We quantify these distinctions in detail in the next section; here, we look at the overall distributions of these metrics for our dataset.
\vspace*{-2mm}
\paragraph{Emotion mean: } On average, novels (when considered as a single speaker) have a mean value of 0.65 for valence, 0.38 for arousal, and 0.52 for dominance. Narration alone has lower mean values of valence and dominance, and about the same for arousal. Character dialogue, on the other hand, has a higher level of valence and dominance compared to narration --- the third quartile for narration is close to the first quartile for dialogue --- while the mean arousal is slightly lower, the spread of values exceeds the range for narration on either end of the distribution.


\paragraph{Variability: }The average variability for novels (as a whole) lies around 0.053 for valence, 0.058 for dominance, and 0.046 for arousal. Narration has consistently higher average values for variability along all three dimensions when compared to character dialogue (nearly double), but the latter again covers a wider range of values.
\paragraph{Peak Distance and Displacement Lengths:} On average, character dialogue has smaller peak distances when compared to narration across the three dimensions (0.016 vs 0.025 for valence). Displacements below the home base for valence have (significantly) higher peaks compared to those above the home base (for both narration and dialogue).
Specifically for character dialogue, the average length of low valence displacements is much higher than all other displacements (106 words, compared to 87 for high valence, and an average of  87 and 82 words for arousal and dominance respectively). This indicates that the most common displacement in the emotional states of characters is when they enter more negative states.

\paragraph{Rise and Recovery Rates: }For narration, rise rates are significantly higher when compared to recovery rates, for all three dimensions, and for both low and high displacements (all metrics are reported in the Appendix Tables \ref{tab:valence-ued-all}-\ref{tab:dom-ued-all}). For character dialogue, we do not see significant ($p < 0.05$) differences for valence (low and high displacements), and for high dominance displacements (i.e., rise and recovery rates are quite similar). 

We note several outliers on either extreme for these metrics, which are potentially interesting to researchers in literary studies, as they pinpoint characters with particularly extreme and notable personality traits. For example, the character of Dr.\@ Watson in the Sherlock Holmes novel \textit{The Sign of the Four} has the highest rise rate for high arousal, indicating that he is easily excited (rises quickly to states of high activity/arousal). We report all meta-speakers and characters that emerge as outliers in Table \ref{tab:vad-mean-std-outliers} in Appendix \ref{appendix-out}.
\subsection{Emotion Arcs Within a Novel (RQ2)}
In the previous section, we compared the aggregate UED metrics for the different text streams in a novel: the entire text, narration, and character dialogue. Here, we instead compare the shapes of the trajectories by considering a measure of arc similarity. We temporally align a pair of arcs that we wish to compare, and compute the Spearman correlation between them (details of the process are in Appendix \ref{sec:appendix-align}). A higher correlation score implies that the two arcs follow similar shapes.

We continue to further quantify the diversity of emotion arcs that can be found within a single story. 

\paragraph{Q1:}\textit{How similar are the emotion arcs of the narration and dialogue in a novel (regardless of which character is uttering it)?}
\\
This question gets at the heart of how close the emotions expressed in the narration are to those of its characters' utterances. In a way, this metric captures a facet of narrative style --- is the narrator emotionally detached from what the characters are experiencing, or does the linguistic style tend to reflect their emotional journeys?


\begin{figure*}
	\centering
	
	\begin{subfigure}{0.3\textwidth}
		\centering
		\includegraphics[width=\textwidth, height=4.1cm]{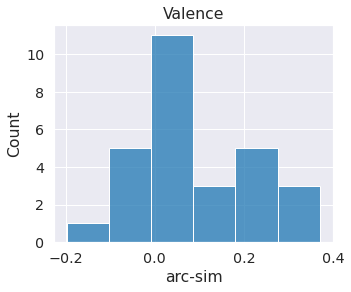}
             \vspace*{-5mm}
		\label{fig:narr-dial-valence-hist}
	\end{subfigure}
	\begin{subfigure}{0.3\textwidth}
		\centering
		\includegraphics[width=\textwidth, height=4.1cm]{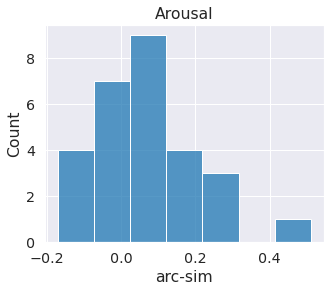}
         \vspace*{-5mm}
		\label{fig:narr-dial-arousal-hist}
	\end{subfigure}
	\begin{subfigure}{0.3\textwidth}
		\centering
		\includegraphics[width=\textwidth, height=4.1cm]{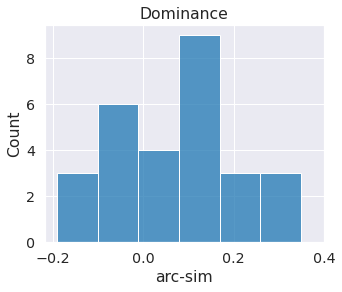}
         \vspace*{-5mm}
		\label{fig:narr-dial-dom-hist}
	\end{subfigure}
 \vspace*{-2mm}
 \caption{Distribution of arc correlations between narration and dialogue (irrespective of character) for all three VAD dimensions.} 
	\label{fig:narr-dial-corr-hist}
 \vspace*{-4mm}
\end{figure*}

\textbf{Results: }Figure \ref{fig:narr-dial-corr-hist} plots the distribution of correlation scores between arcs of narration and dialogue for all 28 novels, for each of the three emotion dimensions. On average, we see near-zero correlations across all dimensions (0.06 for arousal and dominance, 0.09 for valence), with most correlation scores falling between $-$0.1 and 0.35 (mild positive correlation).
We note the following outliers on either side of this range: for valence, \textit{Alice's Adventures in Wonderland} has the lowest correlation ($-$0.2), and \textit{The Sport of the Gods} has the highest correlation (0.37), followed by \textit{The Age of Innocence} (0.33). 
Along the dominance dimension, the arcs in \textit{The Awakening} have a mild positive correlation (0.35); for arousal, \textit{Daisy Miller}, a first-person narrative, is a clear outlier with a correlation of 0.51. The latter outlier is more notable in terms of the absolute value of the correlation (moderate), indicating that the emotional tone of the narration is more in tune with the dialogue. 
\\
\textbf{Discussion: }The correlations between the emotional states of the narration and dialogue in novels are surprisingly low; in most cases, they are near-zero. This distinction has rarely been explored in prior work on the emotional arcs of narratives, which has largely treated the entire text of the novel as presenting a singular flow of emotions \cite{Reagan2016TheEA, Kim2017InvestigatingTR, Fudolig2022ADO}, and demonstrates that narration and character dialogue often represent distinct emotional arcs within a novel. 
\vspace{-2mm}
\paragraph{Q2:} \textit{How similar are the emotion arcs of the narration and each of the major characters in a novel?}
\\
The above results tell us that the overall arcs of narration and dialogue tend to not be correlated. We now ask if, rather than being equally distant from or close to \textit{all} the characters, the narration tends to attach itself the emotional arc of just its protagonist(s) or antagonist(s). For each novel, we measure the similarity between the arc of the narration and each of its \textit{major} characters.

\textbf{Results:} Figure \ref{fig:narr-major-corr-hist} plots the distribution of correlation scores for major characters from all 28 novels, for the valence dimension. Arcs of narration have little to no correlation with those of the major characters (average values of 0.03, 0.02, and 0.03 for valence, arousal, and dominance respectively).\footnote{Since correlation is a bi-polar scale, we also compute average values for positive and negative correlations separately; none of them go beyond 0.18.} Figure \ref{fig:narr-major-box-by-novel} in Appendix \ref{appendix-by-novel} plots the distribution of correlations for each novel.

Interestingly, though 5 out of 28 novels in PDNC are written in the first-person\footnote{\textit{Daisy Miller}, \textit{The Mysterious Affair at Styles}, \textit{The Sun Also Rises}, \textit{The Sign of the Four}, and \textit{The Gambler}.}, many of the correlations between the narration and the dialogue of the narrator fall in the mild ($-$0.1 to 0.2) range for valence; for arousal, \textit{Daisy Miller} (0.37) has one of the highest correlation scores with its narrator, whereas \textit{The Sign of The Four} (-0.36) has among the lowest.
A close reading of these novels might provide critical literary insights into these results. With \textit{The Sun Also Rises}, we have a traumatized and repressed narrator: what he says out loud is vastly different from what he thinks privately. For other novels, these results could be indicative of the difference between the emotions a character has in real time (dialogue) vs.\@ those in retrospect (narration).
\\
\textbf{Discussion: }These findings reinforce the distinctions between the narration arc and those of a novel's main characters --- we average at near-zero correlations for all three emotion dimensions, highlighting that any computational modeling of the emotion arcs of stories should consider the distinctions between these facets of a narrative. 
\subsection{Diversity of Character Arcs (RQ3)}
\begin{figure*}
	\centering
	
	\begin{subfigure}{0.3\textwidth}
		\centering
		\includegraphics[width=\textwidth, height=4.1cm]{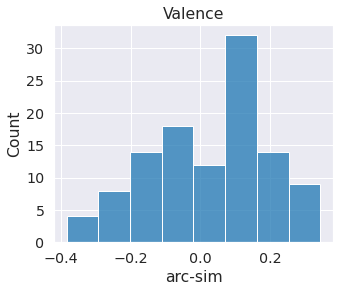}
             \vspace*{-5mm}
		\caption{Between narration and a major character (from the same novel).} 
		\label{fig:narr-major-corr-hist}
	\end{subfigure}
	\begin{subfigure}{0.3\textwidth}
		\centering
		\includegraphics[width=\textwidth, height=4.1cm]{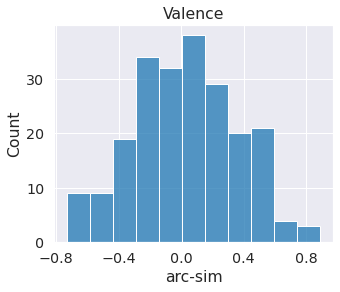}
         \vspace*{-5mm}
		\caption{Between major characters within a novel.}
		\label{fig:major-within-corr-hist}
	\end{subfigure}
	\begin{subfigure}{0.3\textwidth}
		\centering
		\includegraphics[width=\textwidth, height=4.1cm]{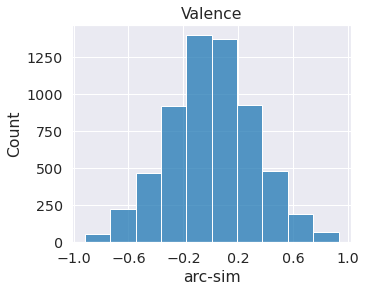}
         \vspace*{-5mm}
		\caption{Between major characters across all novels.}
		\label{fig:major-all-corr-hist}
	\end{subfigure}
 \vspace*{-2mm}
 \caption{Distribution of valence arc correlations.} 
	\label{fig:narr-major-arcs}
 \vspace*{-4mm}
\end{figure*}

We now focus solely on the emotion arcs of character dialogue, and quantify their variation in literary novels. We look at this variation both within a single novel, and when compared across stories. These measures inform us of the diversity of emotional trajectories that a character can follow, and gets closer to the question of the ``basic shapes of stories'' (i.e, of character journeys) that many prior works in NLP have attempted to quantify.  
\vspace{-2mm}
\paragraph{Q3:} \textit{What is the average similarity of character arcs in a novel? Where do we see outliers?}
\\
For each pair of major characters within a novel, we compute the similarity of their emotion arcs. Apart from looking at individual pairs of characters, we compute the mean and variance of the scores for all character pairs within a novel --- which ones have the most diverse emotional journeys for their characters, and the least?
\\
\textbf{Results: }We plot the distribution of all pairwise scores for valence in \ref{fig:major-within-corr-hist}. We can immediately see the much larger spread of the scores here (than for narration and dialogue), 
extending into the range of high correlation and anti-correlation (maximum of 0.89 and minimum of $-$0.73). The mean correlations, however, stay around 0 (0.03, 0.02, 0.04 for the VAD dimensions). 

\begin{figure}
\centering
\includegraphics[width=0.5\textwidth]{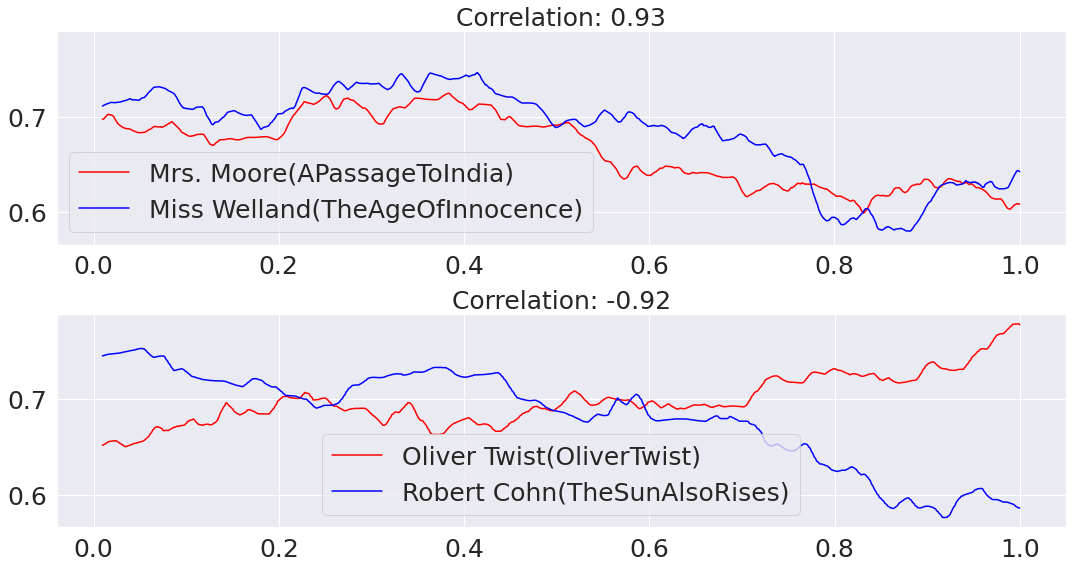}
\caption{Emotion arcs of valence for character pairs with the highest and lowest correlation scores.}
\vspace*{-4mm}
\label{fig:all-major-extreme-arcs}
\end{figure}

We quantify the diversity of character arcs within a novel as the standard deviation of the emotion arc correlations of its major characters. Figure \ref{fig:major-box-by-novel} in Appendix \ref{appendix-by-novel} shows the distribution of correlations for valence by novel. \textit{Winnie-the-Pooh} and \textit{The Sport of the Gods} have some of the highest diversity in major character arcs for all three dimensions; \textit{The Age of Innocence} has the lowest diversity in valence arcs, \textit{The Gambler} and \textit{The Invisible Man} for arousal. 
\textit{The Sun Also Rises} leans the most positive with a median score of 0.44. In this novel, the characters Robert Cohn and Jake Barnes are often read as rivals (they are competing for the love of the same woman), but our analysis shows them travelling the same emotional journey (correlation of 0.89 for valence).
\\
\textbf{Discussion: }The above results lend credence to the variety and diversity that can be found in novels at the level of characters --- it is quite rare to find a story where the  character trajectories or voices are uniform. While it might seem trivial to state that a story is not so much a singular narrative as a collection of intersecting narratives, computational analysis has largely suffered by not being able to afford this complexity in its study of stories. 
The higher correlations between character arcs as opposed to with the narration indicate that 
using a single novel-based arc is a poor representation of the novel for much literary analysis.
\vspace{-2mm}
\paragraph{Q4:} \textit{How similar are the emotion arcs of characters \textit{across} novels?}
\\
This question explores whether we see high correlations reflecting the ``shapes of stories" as described by Kurt Vonnegut, which largely describes the emotional journey of the protagonists of stories. We compute correlation scores between the arcs of all possible pairs of the 111 major characters from all 28 novels (6105 pairs). 
\\
\textbf{Results: }Figure \ref{fig:major-all-corr-hist} plots the distribution of correlation scores for valence (we see similar distributions for arousal and dominance). The average correlation between any pair of characters is close to 0, for all three dimensions. The range of correlation scores is much larger, from a minimum of $-$0.92 (Oliver Twist from \textit{Oliver Twist} and Robert Cohn from \textit{The Sun Also Rises}) to a maximum of 0.93 (Mrs.\@ Moore from \textit{A Passage To India} and Miss Welland from \textit{The Age of Innocence}). In Figure \ref{fig:all-major-extreme-arcs}, we plot the arcs of these two pairs of characters. 
\\
\textbf{Discussion: }The range of the correlations, extending from highly correlated to anti-correlated, demonstrate that character trajectories in different novels can indeed follow similar shapes. However, the Gaussian distribution of the correlation scores and their mean correlations of $\sim$0 
indicate that these arcs do not all follow a small set of prototypical shapes, but cover a wider spectrum.



\subsection{Character Groups (RQ4)}

Several prior works on stories have demonstrated biases in the portrayals of characters that correspond to overgeneralized and often inaccurate stereotypes of certain demographic groups, particularly those that are marginalized \cite{Fast2016ShirtlessAD, sap2017connotation}. We investigate the presence of such biases for the characters in our dataset along the gender dimension, quantifying differences between the UED metrics of Male (M) and Female (F) character groups and how they are written by Male and Female authors. We note that the rather restricted range of novels and authors represented in PDNC \footnote{These are canonically well-regarded authors and novels, who might not be representative of the general fiction landscape of the era, and certainly not of other time-periods.} means we cannot make generalizable claims about these differences; however, our methodology is broadly applicable to any corpus of novels, and provides metrics that are useful to quantify biases in such corpora.

\begin{figure*}
\centering
\includegraphics[width=0.9\textwidth]{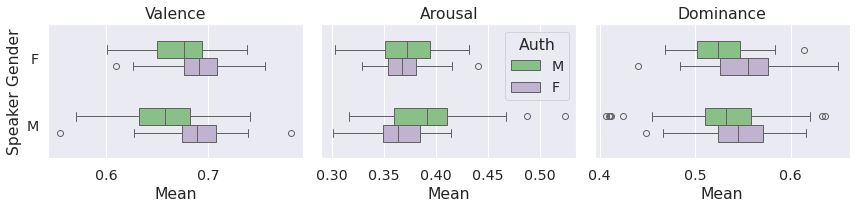}
\caption{Distribution of the emotion mean for VAD, grouped by speaker (y-axis) and author (box color) gender.}
\vspace*{-5mm}
\label{fig:male-female-mean-hist}
\end{figure*}
\vspace{-1mm}
\paragraph{Q5:} \textit{Do the emotional dynamics of characters differ based on their (presented) gender?}
\\
We have 89 Female and 121 Male characters in PDNC with sufficient utterances to compute UED metrics.
We test for statistically significant differences in the aggregate UED metrics (mean, variability, average rise and recovery rates, etc.\@) for each of these groups using a two-sided independent $t$-test, and apply the Benjamini-Hochberg correction for multiple comparisons.
\\
\textbf{Results:} We find the following significant ($p < 0.05$) effects: mean valence is higher for female characters compared to male characters (0.68 vs 0.66); mean arousal is higher for male characters (0.39 vs 0.37), and the average peak distance for arousal displacements is also higher for male characters compared to female characters (0.019 vs 0.016).
\\
\textbf{Discussion: }This is in line with prior work on gender biases in storytelling --- whether in novels, movies, or other forms of stories --- that find that female characters tend to be portrayed with higher levels of positive emotions: warm, kind, caring, and more joyful \cite{Ramakrishna2017LinguisticAO, xu2019cinderella} while male characters are expected to express 
more intense emotions relating to arousal --- anger, rage, and violence.
\vspace{-1mm}
\paragraph{Q6:} \textit{Do male authors write their characters differently from female authors?}
\\
We now compare the UED metrics of characters when grouped by author gender. There are 134 characters written by female authors (85 female, 49 male), and 183 characters written by male authors (133 male, 50 female).
We test for significant differences in aggregate UED metrics for all possible groups of (M/F) characters written by (M/F) authors using the two-way ANOVA test.
\\
\textbf{Results: }Figure \ref{fig:male-female-mean-hist} plots the distributions of mean VAD scores for all non-minor characters in our dataset, separated by speaker and author gender. Character dialogue written by female authors has a higher mean valence (0.69 vs 0.65), lower mean arousal (0.36 vs 0.38), and higher mean dominance (0.55 vs 0.52) when compared to that written by male authors in our dataset ($p < 0.001$). Additionally, we find that the peak distance for low arousal displacements is lower for female-authored characters (0.014 vs 0.019). Male-authored male characters have a particularly high mean arousal when compared to all other groups. 
\\
\textbf{Discussion: }These findings are again in line with prior work on gendered character dialogue in movie scripts and stories \cite{Fast2016ShirtlessAD, xu2019cinderella, Lettieri2023HowMA}, which find that female writers tend to use words that are more positively valenced, and male writers, words that are marked as more arousing. We find significant trends in UED metrics that capture the intensity of emotional displacements, which have not previously been studied. The depiction of the two character gender groups for each of the author groups also reveals trends of interest --- male characters have a higher average arousal in novels authored by male writers, whereas female writers tend to send their male characters into more intense low arousal states.

\section{Conclusion}
In this work, we conducted a detailed analysis of the nuances of the emotional trajectories of stories. We first demonstrated that the emotions expressed in the narration of a story are not representative of those expressed by its characters. We then showed that the characters \textit{within} and \textit{across} stories can have widely varying emotion arcs, with correlations ranging from highly negative to strongly correlated, and averaging at near-zero correlations. These distinctions highlight the diversity of narrative threads contained in a single novel: different characters go through wildly-varying journeys, and these cannot be summarized by a single emotion arc --- of the novel, narration, or dialogue --- alone, or even with a small number 
of prototypical shapes.
\section*{Limitations}
The works that we studied here are set apart by an obvious selection bias, in that the PDNC dataset consists of only 28 novels, chosen because they are popular and critically-acclaimed in the literary canon, written by authors who were societally positioned to be able to achieve fame and success. Will we find more evident, consistent patterns of bias conditioned on character and author gender if we expand our selection to a less-curated, and therefore more representative, dataset of texts?  Perhaps, but automatically extracting information about the number, gender, and types of characters in novels is computationally quite a challenging problem, as is the task of identifying and attributing the various lines of quotation within a novel to one of these characters. We hope progress in these areas of NLP research will enable more expansive analyses of literary characters in the future. 
\bibliography{custom}

\begin{thebibliography}{24}
\expandafter\ifx\csname natexlab\endcsname\relax\def\natexlab#1{#1}\fi

\bibitem[{Baccianella et~al.(2010)Baccianella, Esuli, and
  Sebastiani}]{baccianella-etal-2010-sentiwordnet}
Stefano Baccianella, Andrea Esuli, and Fabrizio Sebastiani. 2010.
\newblock \href
  {http://www.lrec-conf.org/proceedings/lrec2010/pdf/769_Paper.pdf}
  {{S}enti{W}ord{N}et 3.0: An enhanced lexical resource for sentiment analysis
  and opinion mining}.
\newblock In \emph{Proceedings of the Seventh International Conference on
  Language Resources and Evaluation ({LREC}'10)}, Valletta, Malta. European
  Language Resources Association (ELRA).

\bibitem[{Bobicev et~al.(2010)Bobicev, Maxim, Prodan, Burciu, and
  Angheluş}]{Bobicev2010EmotionsIW}
Victoria Bobicev, Victoria Maxim, Tatiana Prodan, Natalia Burciu, and Victoria
  Angheluş. 2010.
\newblock \href {https://api.semanticscholar.org/CorpusID:206743956} {Emotions
  in words: Developing a multilingual {WordNet-Affect}}.
\newblock In \emph{Conference on Intelligent Text Processing and Computational
  Linguistics}.

\bibitem[{Cao and Daum{\'e}~III(2020)}]{cao2020toward}
Yang~Trista Cao and Hal Daum{\'e}~III. 2020.
\newblock Toward gender-inclusive coreference resolution.
\newblock In \emph{Proceedings of the 58th Annual Meeting of the Association
  for Computational Linguistics}, pages 4568--4595.

\bibitem[{Fast et~al.(2016)Fast, Vachovsky, and
  Bernstein}]{Fast2016ShirtlessAD}
Ethan Fast, Tina Vachovsky, and Michael~S. Bernstein. 2016.
\newblock \href {https://api.semanticscholar.org/CorpusID:9293748} {Shirtless
  and dangerous: Quantifying linguistic signals of gender bias in an online
  fiction writing community}.
\newblock \emph{ArXiv}, abs/1603.08832.

\bibitem[{Fudolig et~al.(2022)Fudolig, Alshaabi, Cramer, Danforth, and
  Dodds}]{Fudolig2022ADO}
Mikaela Irene~D. Fudolig, T.~Alshaabi, Kathryn Cramer, Christopher~M. Danforth,
  and Peter~Sheridan Dodds. 2022.
\newblock \href {https://api.semanticscholar.org/CorpusID:258422147} {A
  decomposition of book structure through ousiometric fluctuations in
  cumulative word-time}.
\newblock \emph{Humanities and Social Sciences Communications}, 10:1--12.

\bibitem[{Hipson and Mohammad(2021)}]{hipson2021emotion}
Will~E. Hipson and Saif~M. Mohammad. 2021.
\newblock \href {https://doi.org/10.1371/journal.pone.0256153} {Emotion
  dynamics in movie dialogues}.
\newblock \emph{PLoS ONE}, 16:1--19.

\bibitem[{Hollenstein(2015)}]{Hollenstein2015ThisTI}
Tom Hollenstein. 2015.
\newblock \href {https://api.semanticscholar.org/CorpusID:146967860} {This
  time, it’s real: Affective flexibility, time scales, feedback loops, and
  the regulation of emotion}.
\newblock \emph{Emotion Review}, 7:308 -- 315.

\bibitem[{Hutto and Gilbert(2014)}]{Hutto2014VADERAP}
Clayton~J. Hutto and Eric Gilbert. 2014.
\newblock \href {https://api.semanticscholar.org/CorpusID:12233345} {{VADER}: A
  parsimonious rule-based model for sentiment analysis of social media text}.
\newblock \emph{Proceedings of the International AAAI Conference on Web and
  Social Media}.

\bibitem[{Kim et~al.(2017)Kim, Pad{\'o}, and Klinger}]{Kim2017InvestigatingTR}
Evgeny Kim, Sebastian Pad{\'o}, and Roman Klinger. 2017.
\newblock \href {https://api.semanticscholar.org/CorpusID:7840369}
  {Investigating the relationship between literary genres and emotional plot
  development}.
\newblock In \emph{LaTeCH@ACL}.

\bibitem[{Kuppens and Verduyn(2017)}]{KUPPENS201722}
Peter Kuppens and Philippe Verduyn. 2017.
\newblock \href {https://doi.org/https://doi.org/10.1016/j.copsyc.2017.06.004}
  {Emotion dynamics}.
\newblock \emph{Current Opinion in Psychology}, 17:22--26.
\newblock Emotion.

\bibitem[{Lettieri et~al.(2023)Lettieri, Handjaras, Bucci, Pietrini, and
  Cecchetti}]{Lettieri2023HowMA}
Giada Lettieri, Giacomo Handjaras, Erika Bucci, Pietro Pietrini, and Luca
  Cecchetti. 2023.
\newblock \href {https://api.semanticscholar.org/CorpusID:261578522} {How male
  and female literary authors write about affect across cultures and over
  historical periods}.
\newblock \emph{Affective Science}.

\bibitem[{Mohammad(2012)}]{Mohammad2012FromOU}
Saif~M. Mohammad. 2012.
\newblock \href {https://api.semanticscholar.org/CorpusID:11056955} {From once
  upon a time to happily ever after: Tracking emotions in mail and books}.
\newblock \emph{Decis. Support Syst.}, 53:730--741.

\bibitem[{Mohammad(2018{\natexlab{a}})}]{vad-acl2018}
Saif~M. Mohammad. 2018{\natexlab{a}}.
\newblock Obtaining reliable human ratings of valence, arousal, and dominance
  for 20,000 {E}nglish words.
\newblock In \emph{Proceedings of The Annual Conference of the Association for
  Computational Linguistics (ACL)}, Melbourne, Australia.

\bibitem[{Mohammad(2018{\natexlab{b}})}]{LREC18-AIL}
Saif~M. Mohammad. 2018{\natexlab{b}}.
\newblock Word affect intensities.
\newblock In \emph{Proceedings of the 11th Edition of the Language Resources
  and Evaluation Conference (LREC-2018)}, Miyazaki, Japan.

\bibitem[{Nalisnick and Baird(2013)}]{Nalisnick2013CharactertoCharacterSA}
Eric~T. Nalisnick and Henry~S. Baird. 2013.
\newblock \href {https://api.semanticscholar.org/CorpusID:2113526}
  {Character-to-character sentiment analysis in {S}hakespeare’s plays}.
\newblock In \emph{Annual Meeting of the Association for Computational
  Linguistics}.

\bibitem[{Ramakrishna et~al.(2017)Ramakrishna, Martinez, Malandrakis, Singla,
  and Narayanan}]{Ramakrishna2017LinguisticAO}
Anil Ramakrishna, Victor~R. Martinez, Nikos Malandrakis, Karan Singla, and
  Shrikanth~S. Narayanan. 2017.
\newblock \href {https://api.semanticscholar.org/CorpusID:23770335} {Linguistic
  analysis of differences in portrayal of movie characters}.
\newblock In \emph{Annual Meeting of the Association for Computational
  Linguistics}.

\bibitem[{Reagan et~al.(2016)Reagan, Mitchell, Kiley, Danforth, and
  Dodds}]{Reagan2016TheEA}
Andrew~J. Reagan, Lewis Mitchell, Dilan Kiley, Christopher~M. Danforth, and
  Peter~Sheridan Dodds. 2016.
\newblock \href {https://api.semanticscholar.org/CorpusID:5049787} {The
  emotional arcs of stories are dominated by six basic shapes}.
\newblock \emph{EPJ Data Science}, 5.

\bibitem[{Sap et~al.(2017)Sap, Prasetio, Holtzman, Rashkin, and
  Choi}]{sap2017connotation}
Maarten Sap, Marcella~Cindy Prasetio, Ari Holtzman, Hannah Rashkin, and Yejin
  Choi. 2017.
\newblock \href {https://www.aclweb.org/anthology/D17-1247} {Connotation frames
  of power and agency in modern films}.
\newblock In \emph{EMNLP}.

\bibitem[{Tausczik and Pennebaker(2010)}]{Tausczik2010ThePM}
Yla~R. Tausczik and James~W. Pennebaker. 2010.
\newblock \href {https://api.semanticscholar.org/CorpusID:145665613} {The
  psychological meaning of words: {LIWC} and computerized text analysis
  methods}.
\newblock \emph{Journal of Language and Social Psychology}, 29:24 -- 54.

\bibitem[{Teodorescu and Mohammad(2023)}]{Teodorescu2023EvaluatingEA}
Daniel Teodorescu and Saif~M. Mohammad. 2023.
\newblock \href {https://api.semanticscholar.org/CorpusID:259076341}
  {Evaluating emotion arcs across languages: Bridging the global divide in
  sentiment analysis}.

\bibitem[{Vishnubhotla et~al.(2022)Vishnubhotla, Hammond, and
  Hirst}]{vishnubhotla-etal-2022-project}
Krishnapriya Vishnubhotla, Adam Hammond, and Graeme Hirst. 2022.
\newblock \href {https://aclanthology.org/2022.lrec-1.628} {The {Project
  Dialogism Novel Corpus}: A dataset for quotation attribution in literary
  texts}.
\newblock In \emph{Proceedings of the Thirteenth Language Resources and
  Evaluation Conference}, pages 5838--5848, Marseille, France. European
  Language Resources Association.

\bibitem[{Vishnubhotla and Mohammad(2022)}]{VM2022-TED}
Krishnapriya Vishnubhotla and Saif~M. Mohammad. 2022.
\newblock Tweet emotion dynamics: Emotion word usage in tweets from us and
  canada.
\newblock In \emph{Proceedings of the Thirteenth International Conference on
  Language Resources and Evaluation (LREC 2022)}, Marseille, France.

\bibitem[{Vonnegut(2009)}]{vonnegut2009palm}
Kurt Vonnegut. 2009.
\newblock \emph{Palm Sunday: An Autobiographical Collage}.
\newblock Random House Publishing Group.

\bibitem[{Xu et~al.(2019)Xu, Zhang, Wu, and Wang}]{xu2019cinderella}
Huimin Xu, Zhang Zhang, Lingfei Wu, and Cheng-Jun Wang. 2019.
\newblock The {C}inderella complex: Word embeddings reveal gender stereotypes
  in movies and books.
\newblock \emph{PloS one}, 14(11):e0225385.

\end{thebibliography}

\appendix

\section{Utterance Emotion Dynamics}
\label{appendix-a}

\begin{figure*}
\centering
\includegraphics[width=0.7\textwidth]{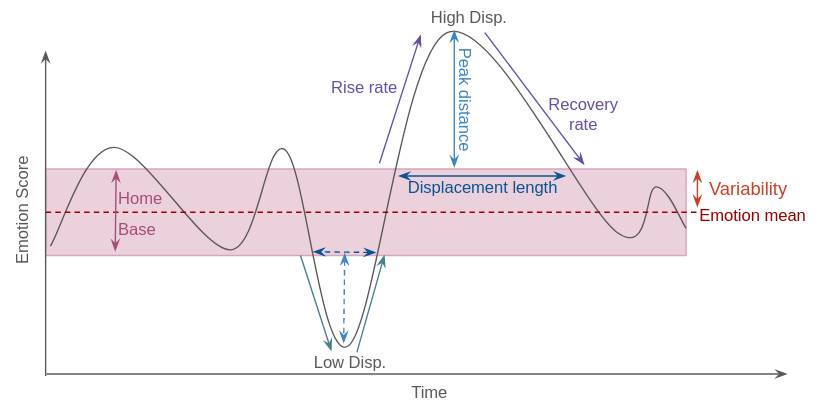}
\caption{A visual representation of the various UED metrics for a sample emotion arc.}
\vspace*{-5mm}
\label{fig:ued-example}
\end{figure*}

The Utterance Emotion Dynamics framework derives several metrics characterizing the temporal patterns of regularity and change of emotion states derived from the textual utterances of an individual. Figure \ref{fig:ued-example} shows a simple example emotion arc and the UED metrics corresponding to the home base and displacements below and above the home base. We briefly describe these metrics here:
\begin{compactitem}
    \item \textbf{Emotion mean (emo\_mean)}: The mean of the sequence of emotion states.
    \item \textbf{Variability (emo\_std)}: The standard deviation of the sequence of emotion states.
    \item \textbf{Average peak distance (emo\_avg\_peak\_dist)}: Average of the peak emotional distance from the home base for all displacements (a measure of how emotional the speaker gets on average when they have a displacement).
    \item \textbf{Average displacement length (emo\_avg\_disp\_length)}: Average of the length of a displacement, in terms of temporal steps, for all displacements (a measure of how long the speaker is outside the home base per displacement on average).
    \item \textbf{Average rise rate (emo\_rise\_rate)}: Average of the rise rates of all displacements (a measure of how quickly one reaches peak distance from home base (regardless of direction of displacement)).
    \item \textbf{Average recovery rate (emo\_recovery\_rate)}: Average of the recovery rates of all displacements (a measure of how quickly one reaches peak distance from home base (regardless of direction of displacement)).
    \item \textbf{Average Low peak distance (emo\_low\_peak\_dist)}: Average of the peak emotional distance from the home base for all displacements below the home base.
    \item \textbf{Average Low displacement length (emo\_low\_disp\_length)}: Average of the length of a displacement, in terms of temporal steps, for all displacements below the home base.
    \item \textbf{Average Home-to-Low rise rate (emo\_low\_rise\_rate)}: Average of the rise rates of all displacements below the home base (measure of how quickly one descends to the lowest emotion state).
    \item \textbf{Average Low-to-Home recovery rate (emo\_low\_recovery\_rate)}: Average of the recovery rates of all displacements below the home base (measure of how quickly one recovers from the lowest emotion state).
    \item \textbf{Average High peak distance (emo\_high\_peak\_dist)}: Average of the peak emotional distance from the home base for all displacements above the home base.
    \item \textbf{Average High displacement length (emo\_high\_disp\_length)}: Average of the length of a displacement, in terms of temporal steps, for all displacements above the home base.
    \item \textbf{Average Home-to-High rise rate (emo\_high\_rise\_rate)}: Average of the rise rates of all displacements above the home base (measure of how quickly one ascends to the lowest emotion state).
    \item \textbf{Average High-to-Home recovery rate (emo\_high\_recovery\_rate)}: Average of the recovery rates of all displacements above the home base (measure of how quickly one recovers from the highest emotion state).
\end{compactitem}

\section{Dataset: PDNC}
\label{appendix-pdnc}
We list details of the novels in the Project Dialogism Novel Corpus in Table \ref{tab:pdnc}. All details were annotated by the dataset creators \cite{vishnubhotla-etal-2022-project} and were made publicly available at \url{https://github.com/Priya22/project-dialogism-novel-corpus}.

\begin{table*}
\centering
\small
    \begin{tabular}{lllll}
\toprule
Novel Title & Narrator Gender & Genre & Author Name & Author Gender \\
\midrule
A Handful Of Dust & U & literary & Evelyn Waugh & M \\
Alice's Adventures in Wonderland & U & children & Lewis Carroll & M \\
Anne Of Green Gables & U & children & Lucy Maud Montgomery & F \\
A Passage to India & U & literary & Edward Morgan Forster & M \\
A Room With A View & U & literary & Edward Morgan Forster & M \\
Daisy Miller & M & literary & Henry James & M \\
Emma & A & literary & Jane Austen & F \\
Hard Times & U & literary & Charles Dickens & M \\
Howards End & M & literary & Edward Morgan Forster & M \\
Mansfield Park & A & literary & Jane Austen & F \\
Night and Day & U & literary & Virginia Woolf & F \\
Northanger Abbey & A & literary & Jane Austen & F \\
Oliver Twist & U & literary & Charles Dickens & M \\
Persuasion & A & literary & Jane Austen & F \\
Pride and Prejudice & A & literary & Jane Austen & F \\
Sense and Sensibility & A & literary & Jane Austen & F \\
The Age of Innocence & U & literary & Edith Wharton & F \\
The Awakening & U & literary & Kate Chopin & F \\
The Gambler & M & literary & Fyodor Mikhailovich Dostoevsky & M \\
The Invisible Man & U & scifi & Herbert George Wells & M \\
The Man Who Was Thursday & U & literary & Gilbert Keith  Chesterton & M \\
The Mysterious Affair At Styles & M & crime & Agatha Christie & F \\
The Picture Of Dorian Gray & U & literary & Oscar Wilde & M \\
The Sign of the Four & M & crime & Arthur Conan Doyle & M \\
The Sport of the Gods & U & literary & Paul Laurence Dunbar & M \\
The Sun Also Rises & M & literary & Ernest Hemingway & M \\
Where Angels Fear to Tread & U & literary & Edward Morgan Forster & M \\
Winnie-The-Pooh & M & children & Alan Alexander Milne & M \\
\bottomrule
\end{tabular}
\caption{Metadata for the 28 novels in PDNC, which we use in our experiments.}
\label{tab:pdnc}
\end{table*}

\section{UED Metrics}
\label{appendix-ued-metrics}
We report the aggregated UED metrics for the different types of speakers -- novel (meta-speaker), narration (meta-speaker), all characters, major characters, intermediate characters, and minor characters -- for valence (Table \ref{tab:valence-ued-all}), arousal (Table \ref{tab:arousal-ued-all}), and dominance (Table \ref{tab:dom-ued-all}).

\begin{table*}
\centering
\small
\begin{tabular}{lrrrrrr}
\toprule
metric // speaker\_type & novel & narration & character & major & intermediate & minor \\
\midrule
emo\_mean & 0.647 & 0.627 & 0.675 & 0.667 & 0.678 & 0.684 \\
emo\_std & 0.053 & 0.057 & 0.029 & 0.038 & 0.026 & 0.018 \\
\hline
emo\_avg\_peak\_dist & 0.023 & 0.025 & 0.016 & 0.019 & 0.015 & 0.012 \\
emo\_avg\_disp\_length & 132.422 & 134.792 & 78.854 & 101.728 & 70.969 & 44.377 \\
emo\_rise\_rate & 0.001 & 0.001 & 0.001 & 0.001 & 0.001 & 0.001 \\
emo\_recovery\_rate & 0.001 & 0.001 & 0.001 & 0.001 & 0.001 & 0.001 \\
\hline
emo\_low\_peak\_dist & 0.027 & 0.028 & 0.021 & 0.025 & 0.020 & 0.012 \\
emo\_low\_disp\_length & 137.656 & 140.427 & 105.963 & 135.199 & 100.196 & 48.943 \\
emo\_low\_rise\_rate & 0.001 & 0.001 & 0.001 & 0.001 & 0.001 & 0.001 \\
emo\_low\_recovery\_rate & 0.001 & 0.001 & 0.001 & 0.001 & 0.001 & 0.001 \\
\hline
emo\_high\_peak\_dist & 0.020 & 0.022 & 0.016 & 0.018 & 0.015 & 0.014 \\
emo\_high\_disp\_length & 130.319 & 136.301 & 87.169 & 110.648 & 76.490 & 55.121 \\
emo\_high\_rise\_rate & 0.001 & 0.001 & 0.001 & 0.001 & 0.001 & 0.002 \\
emo\_high\_recovery\_rate & 0.001 & 0.001 & 0.001 & 0.001 & 0.001 & 0.001 \\
\bottomrule
\end{tabular}
\caption{Averaged UED metrics (rows) of \textbf{valence} for the different speaker types (columns) in PDNC.}
\label{tab:valence-ued-all}
\end{table*}

\begin{table*}
\centering
\small
\begin{tabular}{lrrrrrr}
\toprule
metric // speaker\_type & novel & narration & character & major & intermediate & minor \\
\midrule
emo\_mean & 0.381 & 0.384 & 0.377 & 0.381 & 0.377 & 0.366 \\
emo\_std & 0.047 & 0.046 & 0.028 & 0.037 & 0.024 & 0.016 \\
\hline
emo\_avg\_peak\_dist & 0.020 & 0.021 & 0.016 & 0.021 & 0.014 & 0.011 \\
emo\_avg\_disp\_length & 120.911 & 121.039 & 78.358 & 106.449 & 66.519 & 39.966 \\
emo\_rise\_rate & 0.001 & 0.001 & 0.001 & 0.001 & 0.001 & 0.001 \\
emo\_recovery\_rate & 0.001 & 0.001 & 0.001 & 0.001 & 0.001 & 0.001 \\
\hline
emo\_low\_peak\_dist & 0.018 & 0.018 & 0.017 & 0.020 & 0.015 & 0.012 \\
emo\_low\_disp\_length & 126.509 & 120.441 & 89.166 & 118.320 & 75.781 & 49.623 \\
emo\_low\_rise\_rate & 0.001 & 0.001 & 0.001 & 0.001 & 0.001 & 0.001 \\
emo\_low\_recovery\_rate & 0.001 & 0.001 & 0.001 & 0.001 & 0.001 & 0.001 \\
\hline
emo\_high\_peak\_dist & 0.023 & 0.025 & 0.019 & 0.024 & 0.016 & 0.011 \\
emo\_high\_disp\_length & 118.200 & 125.468 & 85.401 & 117.221 & 71.227 & 40.459 \\
emo\_high\_rise\_rate & 0.001 & 0.001 & 0.001 & 0.001 & 0.001 & 0.001 \\
emo\_high\_recovery\_rate & 0.001 & 0.001 & 0.001 & 0.001 & 0.001 & 0.001 \\
\bottomrule
\end{tabular}
\caption{Averaged UED metrics (rows) of \textbf{arousal} for the different speaker types (columns) in PDNC.}
\label{tab:arousal-ued-all}
\end{table*}

\begin{table*}
    \centering
    \small
    \begin{tabular}{lrrrrrr}
\toprule
metric // speaker\_type & novel & narration & character & major & intermediate & minor \\
\midrule
emo\_mean & 0.519 & 0.508 & 0.538 & 0.532 & 0.545 & 0.536 \\
emo\_std & 0.058 & 0.060 & 0.036 & 0.045 & 0.032 & 0.022 \\
\hline
emo\_avg\_peak\_dist & 0.025 & 0.026 & 0.020 & 0.023 & 0.020 & 0.015 \\
emo\_avg\_disp\_length & 121.263 & 124.919 & 70.747 & 91.681 & 63.142 & 38.689 \\
emo\_rise\_rate & 0.001 & 0.001 & 0.002 & 0.002 & 0.002 & 0.002 \\
emo\_recovery\_rate & 0.001 & 0.001 & 0.001 & 0.001 & 0.001 & 0.002 \\
\hline
emo\_low\_peak\_dist & 0.026 & 0.025 & 0.023 & 0.027 & 0.022 & 0.014 \\
emo\_low\_disp\_length & 124.514 & 126.597 & 86.892 & 116.789 & 74.472 & 39.135 \\
emo\_low\_rise\_rate & 0.001 & 0.001 & 0.002 & 0.001 & 0.002 & 0.002 \\
emo\_low\_recovery\_rate & 0.001 & 0.001 & 0.001 & 0.001 & 0.001 & 0.001 \\
\hline
emo\_high\_peak\_dist & 0.025 & 0.026 & 0.021 & 0.023 & 0.021 & 0.018 \\
emo\_high\_disp\_length & 121.133 & 126.999 & 77.927 & 94.625 & 71.660 & 51.979 \\
emo\_high\_rise\_rate & 0.001 & 0.001 & 0.002 & 0.002 & 0.002 & 0.002 \\
emo\_high\_recovery\_rate & 0.001 & 0.001 & 0.001 & 0.001 & 0.002 & 0.002 \\
\bottomrule
\end{tabular}
\caption{Averaged UED metrics (rows) of \textbf{dominance} for the different speaker types (columns) in PDNC.}
\label{tab:dom-ued-all}
\end{table*}


\section{Outlier Characters}
\label{appendix-out}
Speakers who emerge as outliers for the emotion mean and variability metrics, along all three VAD dimensions, are listed in Table \ref{tab:vad-mean-std-outliers}. Speakers can include meta-speakers (like narration). Outliers are identified as points that fall outside the whiskers of a box-and-whisker plot, i.e, the scores that are below $Q_{1}-1.5*IQR$ (low) and above $Q_{3}+1.5*IQR$ (high), where $Q_{1}$ and $Q_{3}$ are the 25th and 75th percentiles (1st and 3rd quartiles) of the distribution, and $IQR(=Q_{3}-Q_{1})$ is termed the inter-quartile range. Outliers are independently identified for each speaker type (novel, narration, character). 


\begin{table*}
\centering
\small
\begin{tabular}{lllllr}
\toprule
       dim &    metric &       meta & extreme &                                      name &  value \\
\midrule
valence &   emo\_std &  narration &    high &          narrator (ThePictureOfDorianGray) &  0.078 \\
& & & & & \\
   valence &  emo\_mean &  character &     low &                        Monks (OliverTwist) &  0.580 \\
   valence &  emo\_mean &  character &     low &                   Mr. Bumble (OliverTwist) &  0.599 \\
   valence &  emo\_mean &  character &     low &        The Invisible Man (TheInvisibleMan) &  0.578 \\
   valence &  emo\_mean &  character &     low &            Mike Campbell (TheSunAlsoRises) &  0.578 \\
   \hline
   arousal &  emo\_mean &  narration &     low &                   narrator (WinnieThePooh) &  0.334 \\
   & & & & & \\
   arousal &  emo\_mean &  character &    high &  Professor De Worms (TheManWhoWasThursday) &  0.452 \\
   arousal &  emo\_mean &  character &    high &           Joe Hamilton (TheSportOfTheGods) &  0.488 \\
   & & & & & \\
   arousal &   emo\_std &  character &     low &        Jock Grant-Menzies (AHandfulOfDust) &  0.008 \\
   arousal &   emo\_std &  character &     low &              Cassandra Otway (NightAndDay) &  0.013 \\
   arousal &   emo\_std &  character &    high &   Prince Charming (ThePictureOfDorianGray) &  0.061 \\
   arousal &   emo\_std &  character &    high &            Mike Campbell (TheSunAlsoRises) &  0.074 \\
   arousal &   emo\_std &  character &    high &                     Piglet (WinnieThePooh) &  0.072 \\
   \hline
   dominance &  emo\_mean &  narration &     low &                   narrator (WinnieThePooh) &  0.403 \\
   & & & & & \\
 dominance &  emo\_mean &  character &     low &               John Andrew (AHandfulOfDust) &  0.412 \\
 dominance &  emo\_mean &  character &     low &                     Eeyore (WinnieThePooh) &  0.407 \\
 & & & & & \\
 dominance &   emo\_std &  character &     low &           Joe Hamilton (TheSportOfTheGods) &  0.008 \\
 dominance &   emo\_std &  character &    high &                 Mary Musgrove (Persuasion) &  0.075 \\
 dominance &   emo\_std &  character &    high &  Christopher Robin - Story (WinnieThePooh) &  0.080 \\
\bottomrule
\end{tabular}
\caption{Outliers among characters and narration on either extreme for the mean and variability metrics, along all three dimensions.}
\label{tab:vad-mean-std-outliers}
\end{table*}

\section{Aligning Emotion Arcs}
\label{sec:appendix-align}

Let's say we want to compare the computed emotion arcs of a set of speakers $S$:

\begin{itemize}
    \item Find the speaker $s_{sm} \in S$ with the smallest temporal length (i.e, the fewest utterances).
    \item Start with an initial timestep window of [0, 0.001] for this speaker (this initial window size is a hyperparameter we examine later). 
    \item Move forward this window by one word for $s_{sm}$ -- this corresponds to a new bin of timestep values for the speaker $s_{sm}$, say $[0.001, 0.011]$. 
    \item Continue to move forward the window by one word until the end of the speaker's arc; this results in a set of $n$ time bins.
    \item Average the emotion state values at the timesteps contained within each the $n$ time bins to obtain $n$ comparable emotion values for each speaker in $S$.
\end{itemize}
Note that, while the window is moving forward by 1 word for $s_{sm}$, it could be moving forward by $k$ words for a different speaker $s_{j}$ who has a longer volume of utterances. However, it ensures that we have an equal number of bins for each speaker, at approximately the same relative time points (i.e, the 1\% mark, the 1.1\% mark, etc).

Once a pair of emotion arcs has been aligned, we obtain equal-sized sequences of emotion states for the normalized bins, for all the speaker arcs we wish to compare (using a similarity metric like the euclidean distance or spearman correlation).

\begin{figure*}
    \centering
     \begin{subfigure}{\textwidth}
      \centering
      \includegraphics[width=\linewidth]{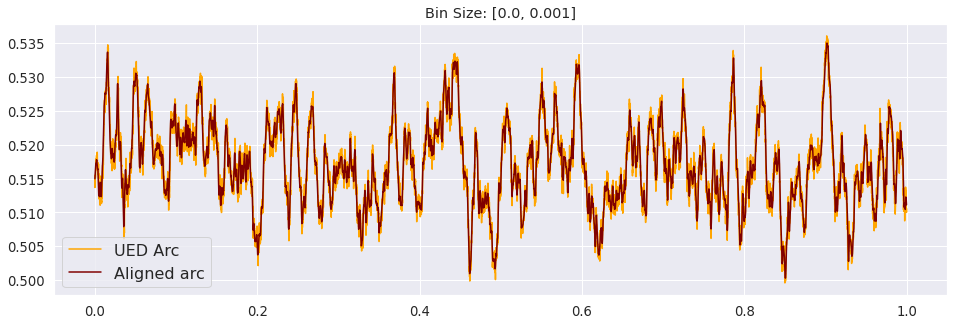}
      \caption{Initial bin size [0, 0.001]}
    \end{subfigure}%
    
     \begin{subfigure}{\textwidth}
      \centering
      \includegraphics[width=\linewidth]{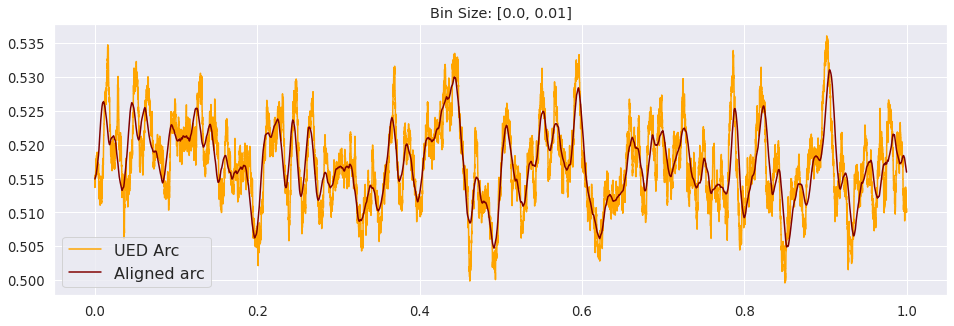}
      \caption{Initial bin size [0, 0.01]}
    \end{subfigure}%
    
     \begin{subfigure}{\textwidth}
      \centering
      \includegraphics[width=\linewidth]{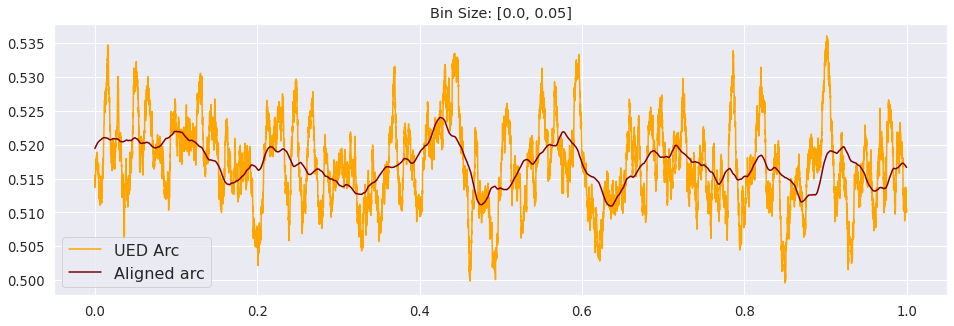}
      \caption{Initial bin size [0, 0.05]}
    \end{subfigure}%
    \caption{A visual comparison of the original UED arc with the time-aligned arc for \textit{A Room With A View} (when aligned with the smallest novel in our dataset, \textit{Daisy Miller}) with different initial bin sizes.}
    \label{fig:smoothing}
\end{figure*}

\textbf{Qualitative Assessment: }How well do the aligned emotion arcs represent the original UED arc of a novel? We qualitatively examine the aligned arcs with three different initial window sizes. The smallest novel out of the set of 28 in PDNC is \textit{Daisy Miller}. With an initial window size of $[0, 0.01]$, we get 22,603 temporal bins; for an initial window of $[0, 0.001]$, we have 22,809 bins; $[0, 0.05]$ yields 21,690 bins.  We therefore compute a new set of aligned arcs for each novel by averaging the emotion state values within each of these temporal bins, for all three temporal bin sizes.

In figure \ref{fig:smoothing}, we plot the original UED arc and the aligned arc for a longer novel, \textit{A Room With A View}, for each of these choices. While a bin size of \([0, 0.001]\) retains many of the sharp transitions of the original arc , and the \([0, 0.05]\) blurs over too many of them, \([0, 0.01]\) seems to be a good compromise in roughly capturing the ups and downs of the emotion arc. We do note, however, that the appropriate bin size is dictated by the research question of interest; if we wanted to obtain a high-level view of the ``shape of a story", we might prefer to smooth the arc even further.

\section{Arc Correlations by Novel}
\begin{figure*}
	\centering
		\begin{subfigure}[b]{0.48\textwidth}
		\centering
		\includegraphics[width=\textwidth]{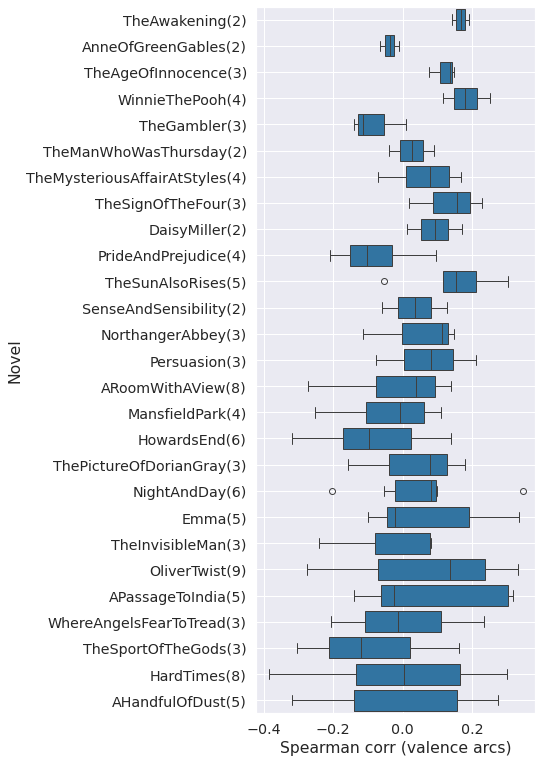}
		\caption{Boxplots of the correlation scores of narration-major character valence arcs for each novel (ordered by variance).}
		\label{fig:narr-major-box-by-novel}
	\end{subfigure}
	\hfill
	\begin{subfigure}[b]{0.48\textwidth}
		\centering
		\includegraphics[width=\textwidth]{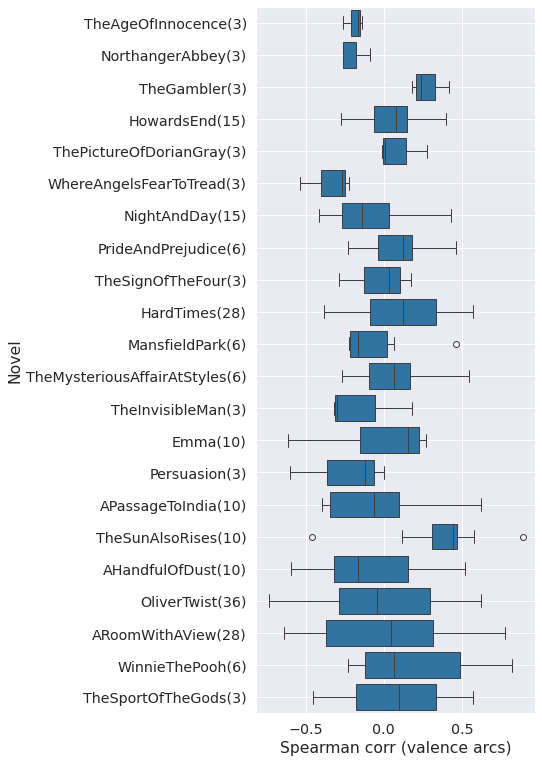}
		\caption{Boxplots of the major character valence arc correlations by novel (ordered by variance).}
		\label{fig:major-box-by-novel}
	\end{subfigure}
	\caption{Within-novel correlations of \textbf{valence} arcs between narration and major character arcs. The numbers in parenthesis beside each novel indicate the number of pairwise correlations that are represented (based on the number of major characters) for each novel.}
	\label{fig:major-arcs}
\end{figure*}

In Figure \ref{fig:narr-major-box-by-novel}, we plot the distributions of the correlations between the arcs of each major character and the narration, for each novel. In Figure \ref{fig:major-box-by-novel}, we report the distributions of the correlations between the arcs of pairs of major characters within each novel.

\label{appendix-by-novel}

\end{document}